\documentclass[letterpaper, 10 pt, conference]{ieeeconf}  

\IEEEoverridecommandlockouts                              

\usepackage{amsmath,amsfonts}
\usepackage{algorithmic}
\usepackage{algorithm}
\usepackage{array}
\usepackage[caption=false,font=normalsize,labelfont=sf,textfont=sf]{subfig}
\usepackage{textcomp}
\usepackage{multirow}
\usepackage{url}
\usepackage{verbatim}
\usepackage{graphicx}
\usepackage{cite}
\usepackage[table]{xcolor} 

\makeatletter
\let\NAT@parse\undefined
\makeatother
\usepackage[pagebackref=true,breaklinks=true,colorlinks,bookmarks=false]{hyperref}

\usepackage[noabbrev,nameinlink,capitalise]{cleveref} 
\Crefformat{figure}{#2Fig.~#1#3}
\Crefmultiformat{figure}{Figs.~#2#1#3}{ and~#2#1#3}{, #2#1#3}{ and~#2#1#3}
\usepackage{float}
\usepackage{booktabs} 
\usepackage{amsmath} 
\usepackage{graphicx} 
\usepackage{multirow} 
\usepackage[table]{xcolor} 
\usepackage{array} 
\title{\LARGE \bf
CogStereo: Neural Stereo Matching with Implicit Spatial Cognition Embedding
}

\author{Lihuang Fang$^{1}$, Xiao Hu$^{2}$, Yuchen Zou$^{3}$, Hong Zhang$^{{1},*}$, \textit{Life Fellow, IEEE}
\thanks{$^{*}$Corresponding author: Hong Zhang}
\thanks{$^{1}$L.Fang and H.Zhang are with the Robotics and Computer Vision (RCV) Laboratory, Southern University of Science and Technology (SUSTech), Shenzhen, China (email: l.h.fang228@gmail.com; hzhang@sustech.edu.cn).}
\thanks{$^{2}$X.Hu is with International Digital Economy Academy~(IDEA), ShenZhen, China (email: huxiao1@idea.edu.cn).}
\thanks{$^{3}$Yuchen Zou is with the School of Automation Science and Engineering, Xi’an Jiaotong University (XJTU), Xi’an, Shaanxi, China (email: yuchenzou@stu.xjtu.edu.cn). }
}
\begin{document}

\maketitle
\thispagestyle{empty}
\pagestyle{empty}

\begin{abstract}

Deep stereo matching has advanced significantly on benchmark datasets through fine-tuning but falls short of the zero-shot generalization seen in foundation models in other vision tasks. We introduce CogStereo, a novel framework that addresses challenging regions, such as occlusions or weak textures, without relying on dataset-specific priors. CogStereo embeds implicit spatial cognition into the refinement process by using monocular depth features as priors, capturing holistic scene understanding beyond local correspondences. This approach ensures structurally coherent disparity estimation, even in areas where geometry alone is inadequate. CogStereo employs a dual-conditional refinement mechanism that combines pixel-wise uncertainty with cognition-guided features for consistent global correction of mismatches. Extensive experiments on Scene Flow, KITTI, Middlebury, ETH3D, EuRoc, and real-world demonstrate that CogStereo not only achieves state-of-the-art results but also excels in cross-domain generalization, shifting stereo vision towards a cognition-driven approach.
\end{abstract}

\section{Introduction}

Stereo matching, essential for estimating depth from binocular images, remains a core challenge in robotics and computer vision~\cite{liu2023real,10500826,tang2019fmd,zuo2019visual,qiu2024mac,chen2024bio}. Despite advancements with synthetic datasets and powerful neural architectures, robust performance in varied real-world environments is elusive. Challenges arise in \emph{ill-posed regions} like occlusions and weak textures, where pixel correspondence is unreliable. Current state-of-the-art (SOTA) methods often rely on domain-specific tuning with techniques such as cost volumes~\cite{xu2023iterative}, recurrent refinements~\cite{jing2023uncertainty,gao2025iterative}, and transformer-based reasoning~\cite{liu2024global}, limiting their generalization to ill-posed regions.

In contrast, {foundation models} for other vision tasks have shown strong \emph{zero-shot generalization}~\cite{hong2024learning}. Models pretrained for classification~\cite{liu2024grounding,yang2025uncertainty}, segmentation~\cite{ren2024grounded}, and monocular depth estimation~\cite{yang2024depth} have demonstrated robust performance on diverse data without domain adaptation. This raises the question: \textit{Can stereo matching adopt a foundation-model approach for improved generalization to ill-posed regions?} This is crucial for applications like autonomous driving~\cite{yang2019drivingstereo} and robotics~\cite{liu2023real}, where consistent performance  over diverse regions is vital.

\begin{figure}[!t]
    \centering
    \includegraphics[width=1\linewidth]{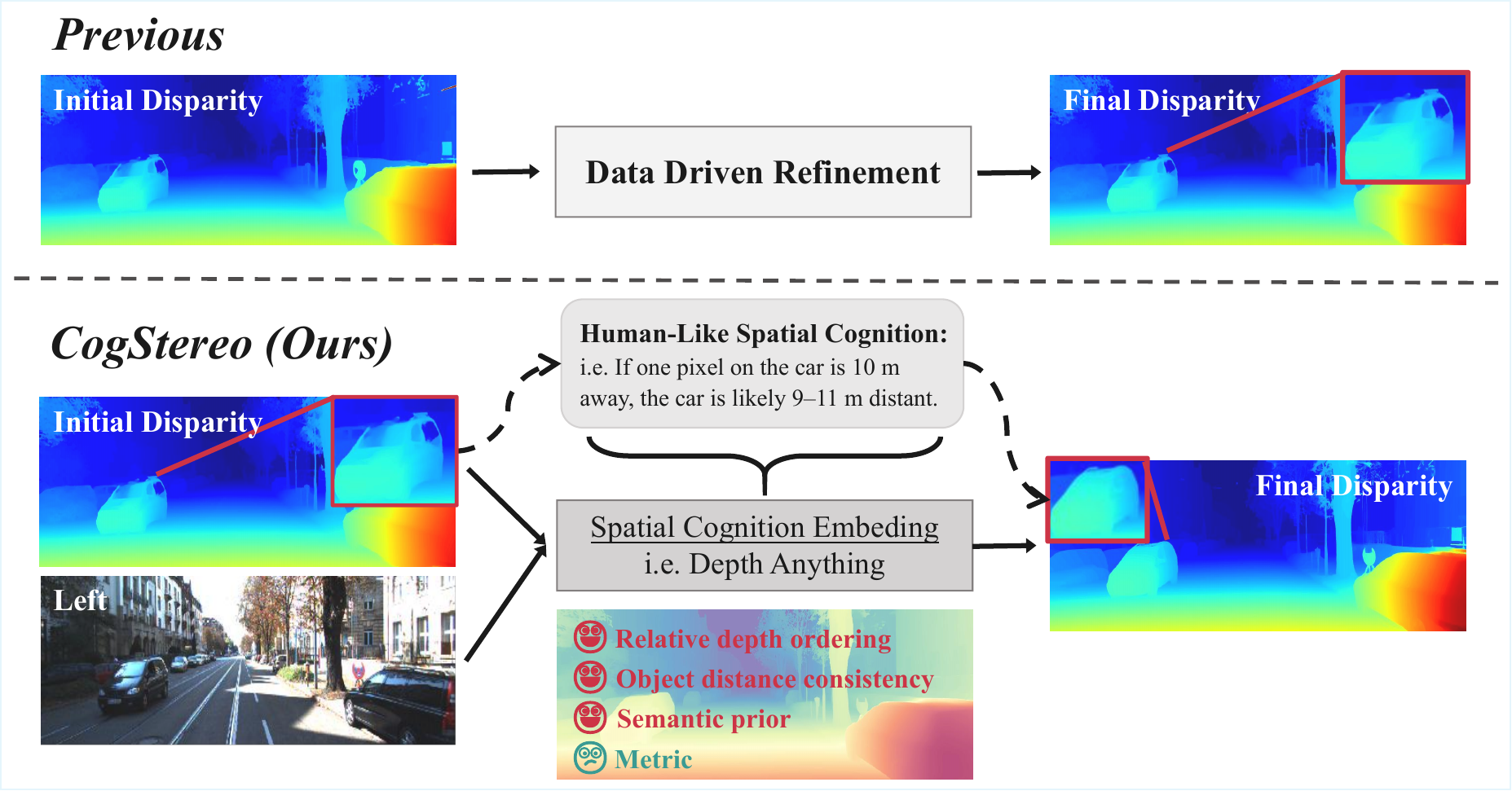}
    \caption{Illustration of the motivation behind CogStereo. While modern stereo networks achieve highly accurate disparity estimation in most regions, they remain vulnerable to mismatches in textureless, reflective, or occluded areas. CogStereo addresses these challenges by embedding implicit SC, enabling globally consistent and robust disparity estimation.
    }
    \label{fig:1}
    \vspace{-0.4cm}
\end{figure}


A key shortcoming in stereo systems is their reliance on local geometric correspondence, often failing in difficult regions. In contrast, the depth foundation model captures object-level geometry and global scene understanding~\cite{yang2024depth}--a capability we term \textbf{\emph{spatial cognition}~(SC)}, similar to human perception (as shown in~\cref{fig:1}). This understanding encompasses depth ordering, shape consistency, and semantic priors (e.g., flat roads, upright humans), allowing monocular models to maintain coherence. Previous attempts to merge monocular cues with stereo~\cite{jiang2025defom} have \emph{explicit depth maps}, which are limited by local errors and lack global consistency.

\begin{figure}[!t]
    \centering\includegraphics[width=1.0\linewidth]{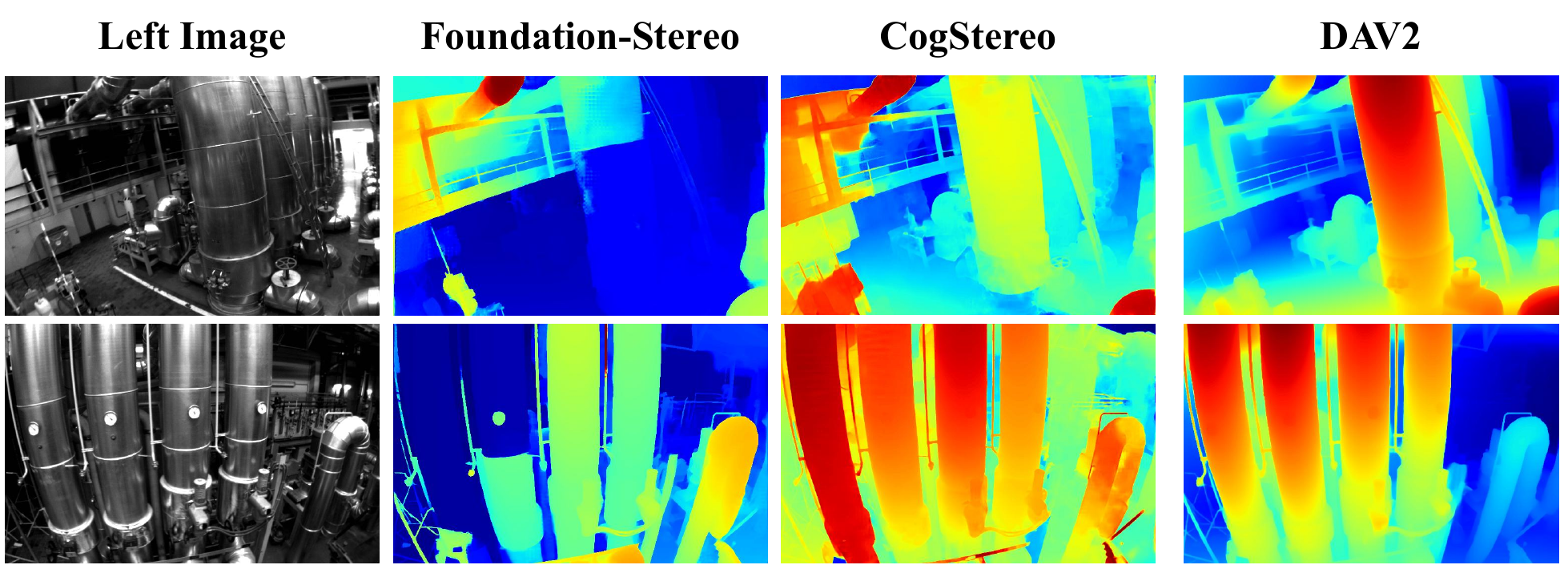}
    \caption{Zero-shot prediction on EuRoC, demonstrating CogStereo's generalization to practical scenarios with challenging properties like weak texture, appearance ambiguities, reflectance, translucency and occlusion.}
    \vspace{-0.5cm}
    \label{fig:vis_foundation}
\end{figure}

To address this, we propose \textbf{CogStereo}, a framework transcending geometric correspondences by embedding implicit SC. Instead of explicit depth predictions, CogStereo utilizes {feature representations} from depth foundation model as {SC priors}. To refine disparities, it introduces a {Dual-Condition Refinement} mechanism, which adjusts disparity based on (i) pixel-wise uncertainty identifiers for unreliable regions and (ii) spatial cognition features for semantic and geometric consistency. This ensures globally coherent disparity maps, even in challenging conditions, as \cref{fig:1} shows how SC aids in achieving robust disparities. 
Overall, \noindent{our contributions are summarized as follows:}
\begin{itemize}
\item We propose \textbf{CogStereo}, embedding \textbf{implicit spatial cognition} into stereo matching, leveraging monocular depth features as priors for improved understanding and accuracy.
\item We introduce a \textbf{Dual-Condition Refinement} mechanism to integrate uncertainty priors with implicit SC features, enhancing disparity correction in ambiguous regions {and preventing metric drift caused by implicit optimization.}
\item Extensive benchmarks and real-world experiments demonstrate CogStereo’s SOTA results and robust zero-shot generalization, marking a shift toward cognitively informed stereo matching.
\end{itemize}
{
\section{Related Work}
\subsection{Learning based Stereo Matching}

Early deep stereo methods~\cite{lipson2021raft,xu2022attention,li2022practical} built 3D cost volumes from binocular features and used 3D convolutions for disparity regression. Later works introduced recurrent refinement~\cite{guo2024lightstereo,lipson2021raft} and transformer-based methods~\cite{li2021revisiting} for long-range pixel correspondences, while NMRF-Stereo~\cite{guan2024neural} used neural MRFs for improved accuracy. Despite these advances, all rely on geometric matching and still exhibit poor generalization in occlusions, weak textures, or repetitive patterns.
Recursive stereo methods like RAFT-Stereo~\cite{lipson2021raft} and its variants improve structure and information integration~\cite{li2022practical,xu2023iterative,wang2024selective}, but still struggle with weak texture or reflective regions. Even large-scale pre-training (FoundationStereo~\cite{wen2025foundationstereo}) shows high zero-shot errors in complex scenes(\cref{fig:vis_foundation}), revealing a cognitive bottleneck in pure stereo paradigms. 
In contrast, CogStereo incorporates spatial cognition from depth foundation model, directly addressing the matching challenge.

\subsection{Monocular Depth Estimation}
Monocular depth estimation has advanced rapidly with deep learning~\cite{bochkovskii2024depth,10552215,yin2023metric3d,hu2024metric3d}. Early CNN-based regression methods suffered from scale ambiguity and domain gaps, while MiDaS~\cite{birkl2023midas} reframed the task as relative scene understanding, enabling zero-shot inference. Depth Anything (DA)~\cite{depthanything} and DAv2~\cite{yang2024depth} further improved robustness through semantic alignment and pseudo-label distillation, and diffusion-based approaches~\cite{10655342,ke2025marigold} leverage generative priors to enforce structural consistency. 
With large-scale pretraining and foundation models such as DINOv2~\cite{oquab2023dinov2}, modern monocular depth models can capture relative depth ordering, object geometry, and global layout. Crucially, they preserve geometric and semantic consistency even in textureless or occluded regions, a property we call \textbf{SC}, reflecting holistic scene understanding and strong zero-shot generalization. This emerging property provides the key inspiration~\cite{lin2025prompting} for our proposed CogStereo, which aims to embed SC into stereo matching for robust and globally consistent disparity estimation.

}

\section{CogStereo}
\label{methodology}
\subsection{Preliminaries}
Given a rectified image pair \(I_l\) and \(I_r\), stereo matching estimates a dense horizontal displacement field \(f_h\), mapping each pixel \((u, v)\) in \(I_l\) to \((u+f_h, v)\) in \(I_r\). A method like RAFT-Stereo~\cite{lipson2021raft} constructs a cost volume by correlating features from the left and right images:
\begin{align}
C(i,j,k) = \frac{F(I_l) \cdot F(I_r)}{\sqrt{D}}, \label{co}
\end{align}
where \(F(\cdot)\) denotes feature maps, \(D\) is the feature dimension, and \(C(i,j,k)\) measures the similarity between pixel \((i,j)\) in \(I_l\) and displacement \(k\) in \(I_r\). Disparity is then obtained by iterative refinement over the cost volume.  
While effective in textured regions, these {matching and refinement methods} often fail in areas such as textureless surfaces, reflections, repetitive patterns, and occlusions where correspondences are ambiguous. The underlying limitation is the absence of {SC}: reasoning is restricted to local feature similarity without object- or scene-level awareness. 

\begin{figure*}[ht]
    \centering
    \includegraphics[width=0.95\linewidth]{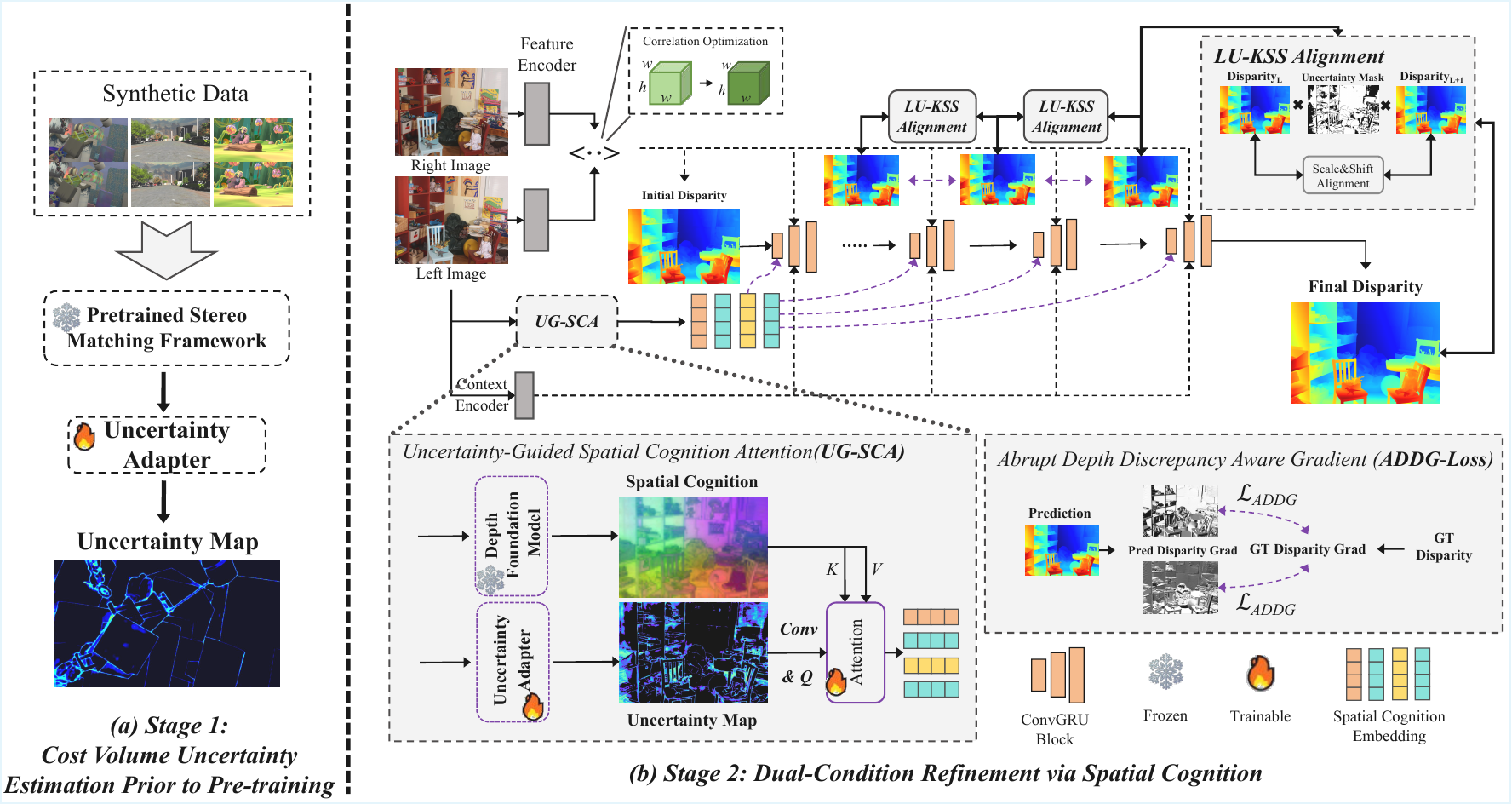}
    \caption{Overview of Our CogStereo Framework. 
    The CogStereo framework enhances stereo matching accuracy through two stages. Stage 1 pretrains the network to generate an uncertainty map from the cost volume. Stage 2 uses the \textit{UG-SCA} module to refine disparities, leveraging SC to correct errors in high-uncertainty regions.
    The ConvGRU Block can be selected from an arbitrary neural stereo matching framework.} 
    \label{fig:overview}
    \vspace{-0.3cm}
\end{figure*}
\vspace{-0.3cm}
\subsection{Framework Overview}
As shown in~\cref{fig:overview}, CogStereo is a novel neural stereo matching method that significantly enhances matching accuracy and robustness through implicit spatial cognition embedding. 
The encoder $F$ from Eq.~\ref{co} first extracts features from $I_l$ and $I_r$ to construct a 3D cost volume, forming the conventional stereo matching backbone.
On top of this baseline, {Our framework utilizes the depth foundation model DAv2\cite{yang2024depth} to extract features rich in spatial cognition, including relative depth ordering, object shape consistency, and physical priors.
Furthermore, to address limitations that metric drift arises when directly applying the DAv2 feature to stereo matching, we introduce several complementary mechanisms. }
During the correlation optimization phase, Uncertainty Adapter is introduced to predict the log-variance of each pixel, which is jointly optimized with disparity regression, allowing uncertainty information to be directly integrated into feature learning and disparity optimization.
In the dual-condition refinement stage via SC, CogStereo integrates pixel-level uncertainty priors with SC features from DAv2.
An uncertainty-guided spatial cognitive attention mechanism identifies areas requiring correction while leveraging reliable SC information to achieve globally consistent refinement.
Meanwhile, to prevent metric drift during implicit optimization, CogStereo employs a KNN-based {{scale-and-shift alignment}} strategy (\textit{LU-KSS}) in low-uncertainty regions. Pixels with uncertainty {below the $\theta$-th percentile serve as anchor points}, and weighted scale and shift are computed via KNN to align sparse reference disparities. Low-uncertainty area are directly anchored, while high-uncertainty region use the average scale and shift alignment. 
In addition, the ADDG-Loss is introduced to penalize abrupt disparity variations in challenging areas.

\subsection{Cost Volume Uncertainty Estimation Prior to Pre-training}  
While modern stereo networks often deliver accurate disparity predictions in most regions, they remain susceptible to errors in textureless surfaces, reflective areas, and occlusions. A key limitation is that existing methods typically estimate uncertainty only after disparity regression, which prevents fully leveraging cost volume information during feature learning and optimization.

To address this, we introduce \textit{Cost Volume Uncertainty Estimation Prior to Pre-training}~(as illustrated in~\cref{fig:overview} ( a )), where uncertainty is explicitly modeled at the cost volume stage. Specifically, we design a lightweight \textit{Uncertainty Adapter} atop the cost volume to jointly predict the log-variance of each pixel alongside disparity regression, thereby embedding confidence information into the optimization process. The adapter consists of two convolutional layers with a ReLU activation in between, outputting a log-variance map that characterizes pixel-wise uncertainty. This uncertainty map is incorporated into the training objective through a residual-adaptive loss formulation:  
\begin{equation}
\mathcal{L}_{\text{uncertainty}} = \tfrac{1}{2} e^{-\log\sigma^2}(d_{{pred}}-d_{{gt}})^2 + \tfrac{1}{2}\log\sigma^2,
\end{equation}
which enforces stronger supervision in confident regions while attenuating gradients in uncertain ones, effectively reducing overfitting and noise propagation.  

\begin{figure}[!t]
    \centering
    \includegraphics[width=0.9\linewidth]{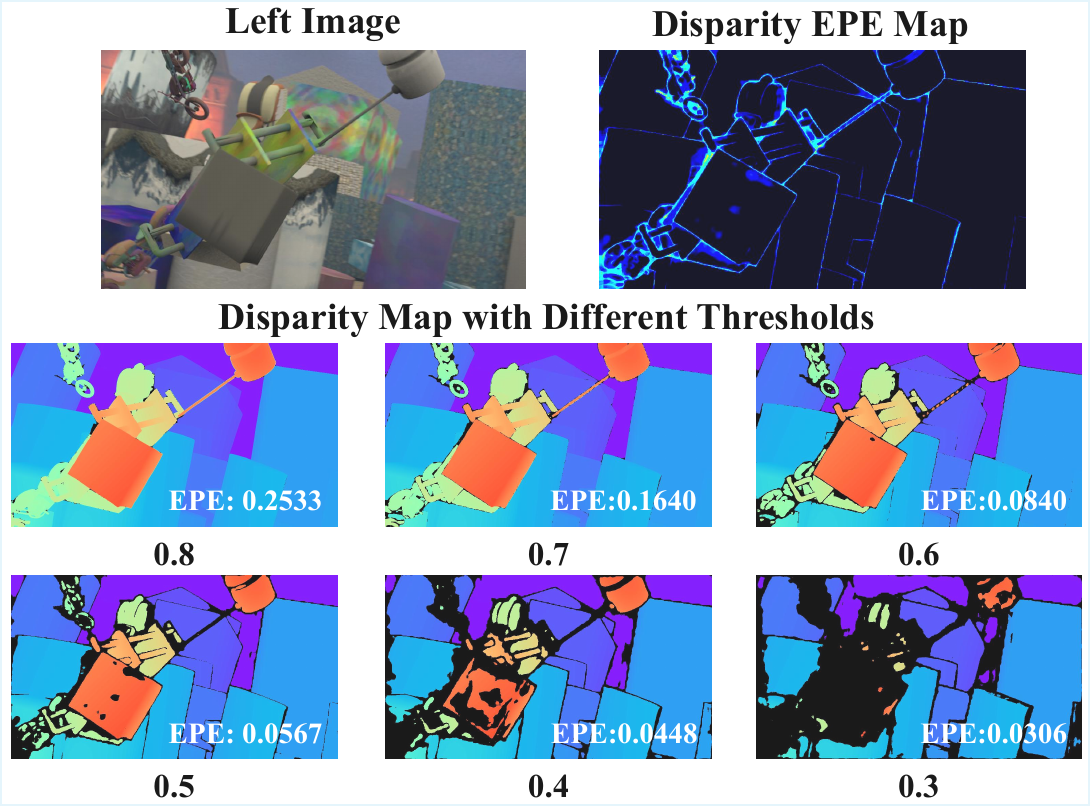}
    \caption{Effectiveness of Uncertainty Masking on Disparity Estimation Accuracy. The top row shows the left image and disparity EPE map, highlighting high-error regions. Subsequent rows show reduced EPE with high uncertainty regions masked, where lower values mean higher accuracy.}
    \label{fig:uncertainty}
    \vspace{-0.5cm}
\end{figure}

By making uncertainty an intrinsic and interpretable signal rather than a post-processing byproduct, our approach produces meaningful uncertainty maps and enables more robust disparity optimization. As shown in~\cref{fig:uncertainty}, thresholding high-uncertainty pixels yields a significant reduction in EPE, highlighting the effectiveness of the proposed uncertainty prior. After learning uncertainty-aware representations, CogStereo leverages them jointly with spatial cognition priors for disparity refinement.
\subsection{Dual-Condition Refinement via Spatial Cognition}

Initial disparity estimates, even from the SOTA stereo or multi-view networks, often contain localized errors due to occlusions, textureless regions, or matching ambiguities. Humans, when assessing depth, instinctively rely on two complementary cues: (i) \textit{error awareness}, signaling unreliable regions, and (ii) \textit{object-level priors}, ensuring consistent depth across the same object. Inspired by this and {condition control\cite{zhang2023adding}}, we propose a \textbf{dual-condition refinement mechanism}~(as illustrated in~\cref{fig:overview} ( b )), where disparity correction is guided by (i) a \textit{pixel-wise uncertainty prior}, indicating \textit{where} corrections are needed, and (ii) \textit{DAv2 features}, offering object-level semantic and geometric cues for \textit{how} to correct errors. This design allows the network to focus on high-uncertainty regions while propagating reliable depth information across object surfaces, yielding disparity maps that are both locally accurate and globally coherent.

\paragraph{Uncertainty as a Reliability Condition}  
A pre-estimated disparity uncertainty map $\mathbf{U} \in \mathbb{R}^{H \times W}$ is incorporated to indicate \textit{where corrections are required}. High-uncertainty pixels represent potential errors, while low-uncertainty pixels serve as reliable anchors. By directing corrections from reliable to uncertain regions, this design prevents over-optimizing already accurate areas and ensures appropriate depth adjustment in ambiguous regions.
\paragraph{Depth Anything Features as a Spatial Cognition Condition}  
We leverage DAv2~\cite{yang2024depth}, a monocular depth model pre-trained on large-scale imagery, to supply object-level semantic and geometric priors.  Instead of using raw depth predictions, we employ the intermediate feature representation $\mathbf{F} \in \mathbb{R}^{H \times W \times C}$, which encodes structural and semantic consistency. These features serve as implicit spatial cognition cues, guiding \textit{how corrections should be propagated} from reliable to uncertain regions, enabling globally coherent refinement even in challenging scenarios.
\begin{figure*}[!htbp]
    \centering
    \includegraphics[width=1\linewidth]{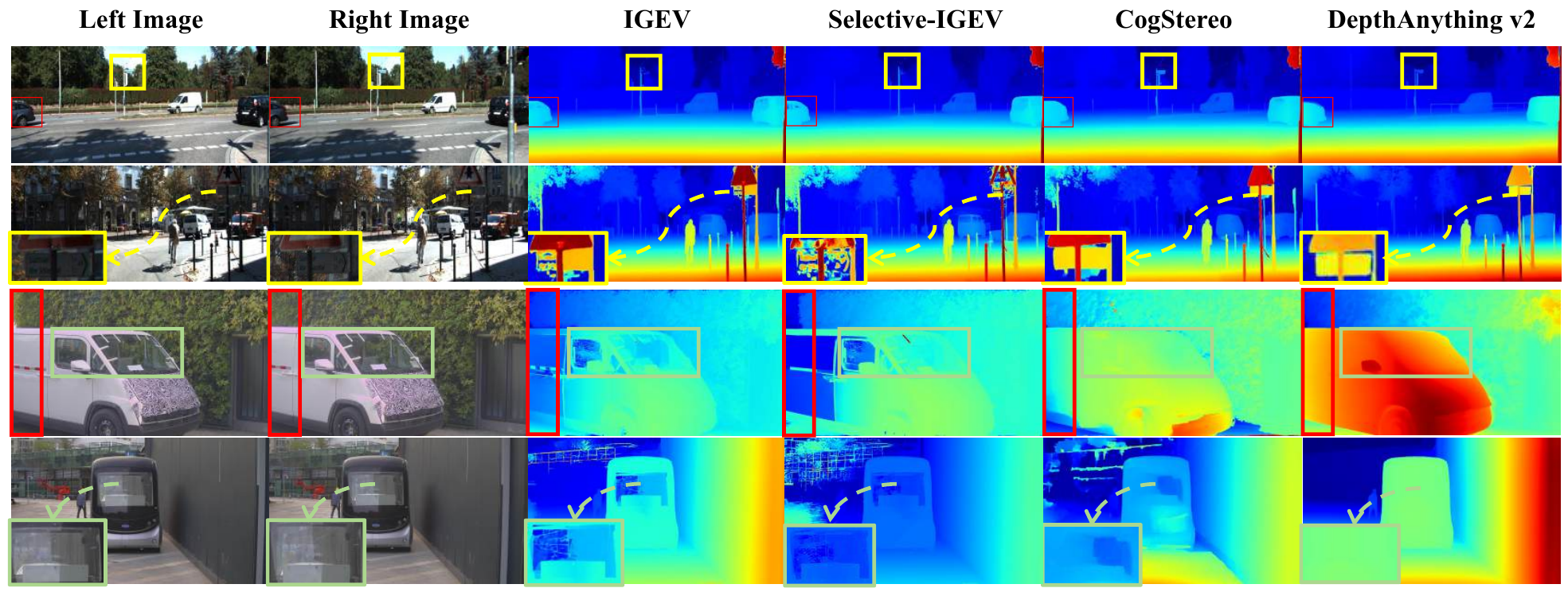}
    \caption{Qualitative comparison of zero-shot inference on in-the-wild images, including samples from the KITTI training dataset and our self-collected mid-range stereo matching dataset for autonomous driving. The \textbf{\textcolor{red}{red boxes}} highlight \textbf{occlusion regions}, the \textbf{\textcolor{green}{green boxes}} indicate \textbf{reflection regions}, and the \textbf{\textcolor{yellow}{yellow }} mark \textbf{weak textures}. CogStereo outperforms the baselines, demonstrating its effectiveness in handling such complexities.}
\label{zeroshotv}
\vspace{-0.3cm}
\end{figure*}
\paragraph{Uncertainty Guided Spatial Cognition Attention( UG-SCA )}
We propose a UG-SCA module that integrates uncertainty information into both spatial cognition and disparity refinement. Specifically, given uncertainty $\mathbf{U}$ and DAv2 features $\mathbf{F}$, uncertainty-conditioned queries interact with features to produce corrective signals $\mathbf{F}_{sc}$:
\begin{equation}
\mathbf{F}_{sc} = \text{Softmax}\Big(\frac{\phi_q(Conv(\mathbf{U})) \cdot \phi_k(\mathbf{F})^\top}{\sqrt{d}}\Big)\phi_v(\mathbf{F}),
\end{equation}
where $\phi_q, \phi_k, \phi_v$ are learnable projections and $d$ is the feature dimension.

\paragraph{Low-Uncertainty Area KNN-Based Scale-and-Shift Alignment (\textit{LU-KSS})}  
To prevent metric drift caused by DAv2 implicit optimization, we propose a KNN-based scale-and-shift alignment strategy guided by low-uncertainty area (LU-KSS), inspired by~\cite{wang2025depth}. First, pixels with uncertainty below the $\theta$-th percentile are selected as reliable anchors:
\begin{equation}
P_{\mathrm{reliable}} = \{ (x,y) \mid U(x,y) \leq \tau_{\theta} \}, 
\quad \tau_{\theta} = Q_{\theta}(U),
\end{equation}
where $Q_{\theta}(U)$ denotes the $\theta$-th percentile of the uncertainty distribution. Next, for each low uncertainty pixel, its $K$ nearest neighbors in the low uncertainty regions are identified, and local alignment parameters $(s, t)$ are estimated via inverse-distance weighted least squares:
\begin{equation}
(s, t) = \arg\min_{s, t} 
\sum_{i=1}^{K} w_i \, \big\| s \, d_{{p,k+1}}(x, y) + t - d_{{p,k}}(x, y) \big\|^2,
\end{equation}
where $d_{{p,k}}(x, y)$ denotes $k$-th predicted disparity and $w_i$ denotes the normalized inverse-distance weight of the $i$-th neighbor. The aligned disparity $d_{{p,k+1}}^{a}$ for low-uncertainty regions is then given by:
\begin{equation}
d_{{p,k+1}}^{a}(x, y) = s \cdot d_{{p,k+1}}(x, y) + t, 
\quad (x,y) \in P_{\mathrm{reliable}},
\end{equation}
For high-uncertainty regions, they do not directly participate in the KNN fitting. Instead, the alignment parameters $(s^\ast, t^\ast)$ for these regions are obtained by averaging the local alignment results from multiple low-uncertainty anchors:
\begin{equation}
(s^\ast, t^\ast) = \frac{1}{|\mathcal{A}|} \sum_{j \in \mathcal{A}} (s_j, t_j),
\quad \mathcal{A} \subseteq P_{\mathrm{reliable}}
\vspace{-0.2cm}
\end{equation}
and the aligned disparity for high-uncertainty regions is computed as:
\begin{equation}
d_{{p,k+1}}^{a}(x, y) = s^\ast \, d_{{p,k+1}}(x, y) + t^\ast, \, (x,y) \in P_{\mathrm{unreliable}},
\end{equation}

\paragraph{Abrupt Depth Discrepancy Aware Gradient Loss}
To improve spatial consistency and reduce abrupt depth changes~\cite{bochkovskiydepth} in disparity estimation, we introduce the \textit{Abrupt Depth Discrepancy Aware Gradient Loss} (ADDG). This loss is specifically designed to penalize sudden changes in disparity, enhancing the smoothness and accuracy of disparity maps. It is particularly effective in regions with low texture, high reflectivity, or occlusions, where traditional stereo methods often struggle. The ADDG loss is defined as:
\begin{equation}
\mathcal{L}_{ADDG} = \left( \left|\frac{\partial (d_{p,k}^a - d_{gt})}{\partial x}\right| + \left|\frac{\partial (d_{p,k}^a - d_{gt})}{\partial y}\right|\right),
\end{equation}
where $d_{p,k}^a$ and $d_{gt}$ denote the aligned predicted and ground-truth disparity, respectively, and $\frac{\partial (d_{p,k}^a - d_{gt})}{\partial x}$, $\frac{\partial (d_{p,k}^a - d_{gt})}{\partial y}$ are the gradients of their difference along the $x$ and $y$ directions.

\subsection{Learning Objectives}
The learning ojectives for the CogStereo is defined as:
\begin{equation}
\vspace{-0.2cm}
\begin{split}
\mathcal{L}_{{total}} = \mathcal{L}_{{init}} + \sum_{k=1}^{N} \gamma^{N-k} & \| d_{p,k}^a - d_{{gt}} \|_1 \\ 
& + \mathcal{L}_{ADDG} + \mathcal{L}_{{uncertainty}}
\end{split}
\end{equation}
where $\mathcal{L}_{{init}} = {L_{Smooth}}(d_{p,0} - d_{{gt}})$ denotes the smooth $\mathcal{L}_{1}$ loss of the initial disparity $d_{p,0}$, $\| d_{p,k}^a - d_{{gt}} \|_1$ is the $\mathcal{L}_{1}$ loss between the aligned disparity $d_{p,k}$ after the $k$-th update and the ground truth disparity $d_{gt}$, $\gamma$ is a decay coefficient, typically set to 0.9, $N$ is the number of iterative updates.

\section{Experiment}
\subsection{Implementation Details}
We implement CogStereo using PyTorch on NVIDIA A100 GPUs, utilizing the AdamW~\cite{loshchilov2017decoupled} optimizer with a one-cycle learning rate scheduler for all experiments. The frozen ViT-L version of DAv2~\cite{yang2024depth} is used to extract spatial cognition embeddings, preserving its pretrained generalization on real-world data. For stereo feature extraction, we utilize the BasicEncoder from IGEV~\cite{xu2023iterative}. Pre-training is conducted on the Scene Flow~\cite{mayer2016large} dataset for 200K iterations with a batch size of 8, using a cosine one-cycle learning rate schedule peaking at 2e-4.
\subsection{Experiment Setting}
\paragraph{Datasets}
We evaluate on five standard benchmarks spanning synthetic and real-world scenarios. Scene Flow \cite{mayer2016large} provides 35,454 synthetic training and 4,370 testing stereo pairs with dense, accurate disparity maps across three subsets: FlyingThings3D, Driving, and Monkaa. For real-world evaluation, KITTI 2012 \cite{geiger2012we} (194 training/195 testing pairs) and KITTI 2015 \cite{geiger2013vision} (200 training/200 testing pairs) offer outdoor urban scenes with sparse LiDAR-based disparity annotations. 
To assess generalization, we additionally include zero-shot evaluation on Middlebury 2014 \cite{scharstein2014high}, featuring high-precision structured-light disparity for indoor scenes, ETH3D \cite{schops2017multi}, containing mixed indoor/outdoor grayscale stereo pairs, EuRoC~\cite{Burri25012016}, and our self-collected mid-range stereo matching dataset for autonomous driving. 
{
\paragraph{Competing Methods}
We compare CogStereo with 11 state-of-the-art~(SOTA) \textbf{data-driven} stereo matching methods that span a broad architectural spectrum. These include classical CNN-based pipelines such as ACVNet~\cite{xu2022attention}, Mask-CFNet~\cite{rao2023masked} RAFT-Stereo~\cite{lipson2021raft}, and PCW-Net~\cite{shen2022pcw}; CNN-Transformer hybrid designs like CREStereo++~\cite{jing2023uncertainty}, IGEV~\cite{xu2023iterative}, IGEV++~\cite{xu2025igev++}, Selective-IGEV~\cite{wang2024selective}, and NMRF-Stereo~\cite{guan2024neural}; as well as the latest Depth-Anything-v2-powered baselines, FoundationStereo~\cite{wen2025foundationstereo} and DEFOMStereo-L~\cite{jiang2025defom}. All methods were trained exclusively on Scene Flow, ensuring a strictly zero-shot and fair cross-dataset evaluation.
\begin{figure*}[htbp]
    \centering
    \includegraphics[width=1\linewidth]{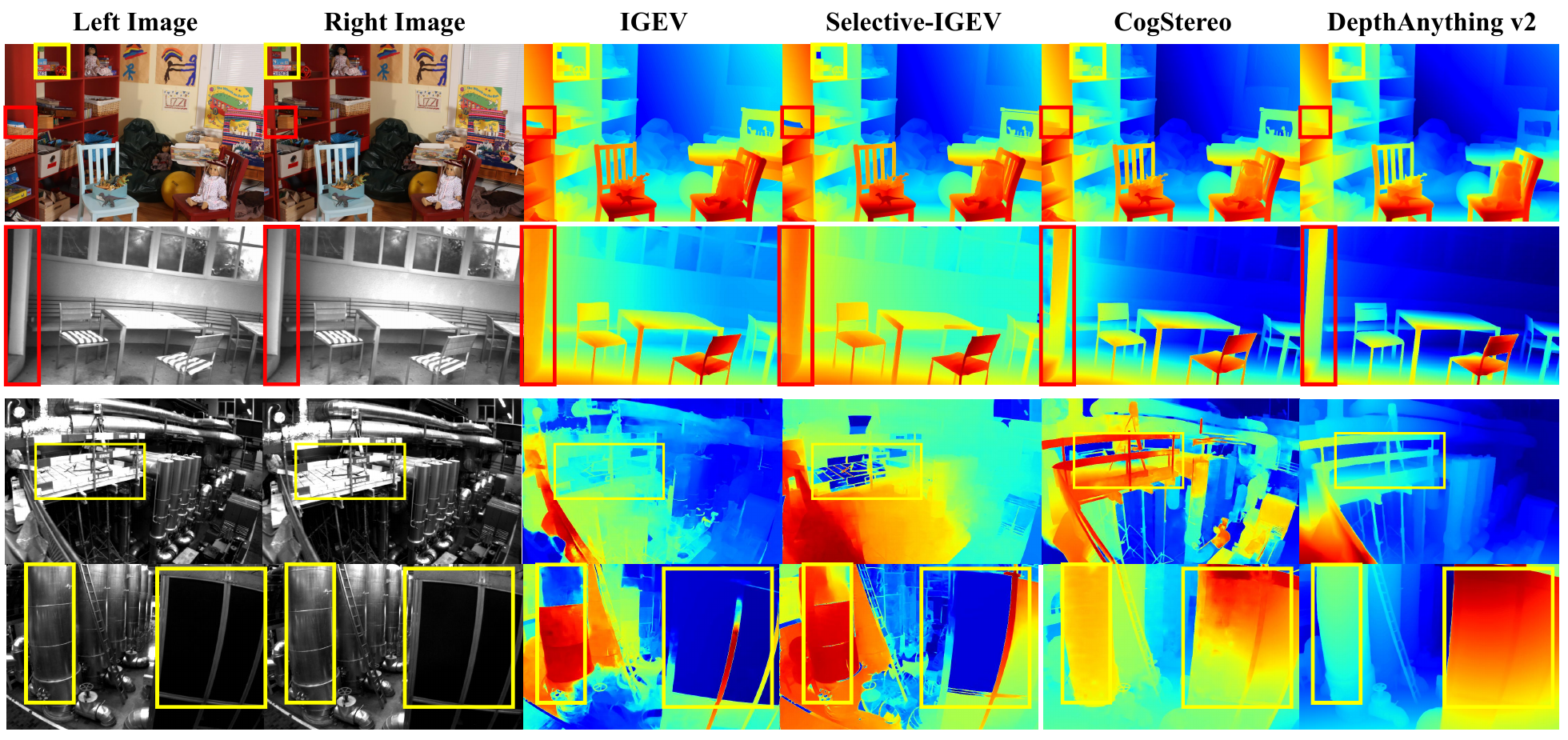}
    \caption{Comparative zero-shot performance of stereo vision algorithms on the Middlebury (1st row), ETH3D (2nd row), EuRoC (last two rows), specifically tailored for Stereo SLAM applications.}    \label{fig:euroc}
    \vspace{-0.2cm}
\end{figure*}

\begin{table*}[!ht]
\centering
\caption{Zero-shot generalization on four public benchmarks. For each dataset, the standard evaluation metrics are reported. All methods are trained exclusively on Scene Flow for a fair comparison. * indicates methods utilizing Depth Anything. NOC, OCC, and ALL denote non-occluded, occluded, and all pixels, respectively. \textbf{\color{green}Green} denotes the best result and {\color{yellow}Yellow} indicates the second best result.}
\resizebox*{0.9\linewidth}{!}{
\label{generalization}
\rowcolors{3}{gray!10}{white} 
\begin{tabular}{l||ccc|ccc|ccc|ccc}
\toprule[1pt]
\multirow{2}*{Methods} & \multicolumn{3}{c|}{\textbf{Middlebury BP-2} $\downarrow$} & \multicolumn{3}{c|}{\textbf{ETH3D BP-1}$\downarrow$} &  \multicolumn{3}{c|}{\textbf{KITTI-12 D1}$\downarrow$} & \multicolumn{3}{c}{\textbf{KITTI-15 D1}$\downarrow$} \\ 
&NOC & OCC & ALL&NOC & OCC & ALL&NOC & OCC & ALL &NOC & OCC & ALL\\
\midrule
ACVNet~\cite{xu2022attention} &22.1& 47.4 & 25.7&8.7& 19.6 & 9.2&  12.9&54.5&  13.9& 11.3&32.9& 11.7 \\
Mask-CFNet \cite{rao2023masked}& - & - & 13.7 & - & -& 5.7& - & - & 4.8 & - & -& 5.8 \\
RAFT-Stereo~\cite{lipson2021raft}&9.1&28.0 &12.0  &2.9&6.0  &3.0 & 4.3&  28.4   & 4.7 & 5.3& 12.7  & 5.5 \\
PCW-Net~\cite{shen2022pcw} &  12.2 &38.0 &15.9 &  5.3&11.7 & 5.5 & 4.1 & 30.2& 4.7& 5.5 & 15.0 &5.7 \\
CREStereo++ \cite{jing2023uncertainty} & - & -& 14.8 & - & -& 4.4 & - & - & 4.7 & - & -& 5.2 \\
IGEV~\cite{xu2023iterative} &7.3&24.3 &9.9&  4.1&9.8&  4.4 & 4.9 & 33.7& 5.6& 5.6 & 14.3 & 5.8 \\
IGEV++ \cite{xu2025igev++} & - & -& 7.8 & - & -& 4.1 & - & -& 5.1 & - & -& 5.9 \\
NMRF-Stereo~\cite{guan2024neural} & - & -& 7.5 & - & -& 3.8 & - & -& 4.2 & - & -& 5.1 \\
Selective-IGEV~\cite{wang2024selective} &6.7&22.6&9.2 &4.1&9.8 &4.4 & 5.1& 31.9 & 5.7 &5.7&13.8& 5.9 \\ 
FoundationStereo*~\cite{wen2025foundationstereo} & - & -& \cellcolor{yellow!50}{5.5}& - & - & \cellcolor{yellow!50}{1.8} & - & -& \cellcolor{yellow!50}{3.2} & - & -& \cellcolor{yellow!50}{4.9} \\
DEFOMStereo-L*~\cite{jiang2025defom} &\cellcolor{yellow!50}{4.4}&\cellcolor{yellow!50}{20.6}& 6.9  &\cellcolor{yellow!50}{2.1}&\cellcolor{yellow!50}{5.1}  &2.2 & \cellcolor{yellow!50}{3.8} & \cellcolor{yellow!50}{22.0}& {4.2} & \cellcolor{yellow!50}{4.8} &\cellcolor{yellow!50}{12.6} & 5.0 \\
\midrule
\textbf{CogStereo}~(Ours) & \cellcolor{green!30}{4.2} & \cellcolor{green!30}{10.0}& \cellcolor{green!30}{5.1} & \cellcolor{green!30}{1.5} & \cellcolor{green!30}{4.0}& \cellcolor{green!30}{1.6} & \cellcolor{green!30}{2.7} & \cellcolor{green!30}{16.3}& \cellcolor{green!30}{3.0}  & \cellcolor{green!30}{4.2} & \cellcolor{green!30}{9.1}& \cellcolor{green!30}{4.3} \\
\bottomrule[1pt]
\end{tabular}
}
\end{table*}

\begin{table*}[ht]
\centering
\caption{Quantitative Evaluation on Scene Flow test set. \textbf{Bold} denotes the best result and \underline{underline} indicates the result of the baseline. {\color{red}Red} denotes the improvement comparison with the baseline.} 
\label{tab:sceneflow}
\resizebox*{0.95\linewidth}{!}{
\begin{tabular}{lcccccc|c}
\toprule
Method & GwcNet \cite{guo2019group} & LEAStereo \cite{cheng2020hierarchical} & RAFT-Stereo~\cite{lipson2021raft} & IGEV \cite{xu2023iterative} & Selective-IGEV \cite{wang2024selective} & DEFOMStereo-L~\cite{jiang2025defom}  & \textbf{CogStereo} \\
\midrule
EPE (px) & 0.76 & 0.78& 0.67& \underline{0.47} & {0.44} & {0.42} & \textbf{0.35}{\color{red}$_{-25.53\%}$}\\
\bottomrule
\end{tabular}
}
\vspace{-0.3cm}
\end{table*}
\paragraph{Evaluation Metric}
Our central hypothesis is that embedding SC from DAv2 enables stereo networks to handle ill-posed regions such as occlusions, textureless, and reflective surfaces, where geometric correspondence alone often fails. To verify this, we adopt three standard metrics: the average end-point error (EPE), the Bad Pixel rate (BP-X) that measures the fraction of pixels with disparity error exceeding $X$, and the D1 metric that considers errors larger than both 3 pixels and 5\% of ground truth. 
Since large errors predominantly occur in ill-posed regions such as occlusions (OCC), weak textures, and reflective surfaces, improvements in BP-X and D1 directly validate CogStereo’s strength in handling these challenges.
\subsection{Zero Shot Quantitative Comparison}
As shown in \cref{zeroshotv,fig:euroc}, we evaluate zero-shot generalization capabilities in real-world scenes featuring \textcolor{red}{occlusions}, \textbf{\textcolor{green}{specular reflections}}, \textcolor{yellow}{weak textures}, and fine-grained structures. Despite lacking a metric scale, the monocular depth model DAv2 generates structurally and semantically consistent depth maps in these challenging regions by leveraging pre-trained SC priors: it outputs correct depths for reflective areas unaffected by objects behind reconstructed glass, maintains consistency within repeating patterns, and connects slender structures. In contrast, the geometry-only IGEV model exhibits noise, discontinuities, and detail loss. By embedding DAv2's structural prior into the IGEV backbone (see \cref{methodology}), CogStereo inherits DAv2's smooth, coherent structures while preserving metric accuracy, enabling robust disparity estimation under zero-shot conditions.
\subsection{Qualitative Analysis}
\paragraph{{Zero Shot Generalization Comparison}}
As shown in~\cref{generalization}, CogStereo demonstrates strong zero-shot generalization across four public benchmarks. Beyond reducing overall errors (ALL), CogStereo consistently improves performance in both {NOC} and {OCC} regions. Notably, in the more challenging {OCC} regions, it achieves 10.0 on Middlebury, 4.0 on ETH3D, 16.3 on KITTI-12, and 9.1 on KITTI-15, surpassing prior SOTA methods by a clear margin. It confirms that spatial cognition priors are particularly beneficial in ill-posed settings such as occlusions, weak textures, and reflective surfaces, enabling CogStereo to recover disparities that are often lost by geometry-only methods. These results highlight that CogStereo’s robust improvements in challenging regions are not coincidental, but stem from the integration of spatial cognition, establishing it as a SOTA framework for zero-shot stereo matching.
\paragraph{{In-Domain Comparison}}
\cref{tab:sceneflow} shows a quantitative comparison on Scene Flow, using the official train-test split. CogStereo surpasses other methods, reducing the best previous EPE from 0.47 to 0.35. Although in-domain training is not the main focus, these results highlight the effectiveness of our model design.
\subsection{Ablation Study}
\paragraph{Ablation Study of CogStereo Module}
We conducted an ablation study on the Scene Flow test set to assess each component's effectiveness, summarized in~\cref{tab:extension1}. The baseline achieves an EPE of 0.47. Removing UG-SCA results in an EPE of 0.44; although slightly improved, the absence of uncertainty guidance causes over-optimization in low-uncertainty regions and under-correction in high-uncertainty areas. Removing LU-KSS worsens the EPE to 0.49, worse than the baseline, due to metric drift without scale–shift alignment. Excluding $\mathcal{L}_{ADDG}$ worsens the EPE to 0.36, showing that penalizing abrupt depth discrepancies enhances robustness in challenging areas. Incorporating all components yields the best EPE of 0.35, confirming that each module contributes to performance improvement.
\begin{table}[!h]
\centering
\caption{Ablation study of CogStereo Module. \textit{w/o} denotes without.}
\label{tab:extension1}
\resizebox*{0.8\linewidth}{!}{
    \renewcommand{\arraystretch}{1.0}
    \rowcolors{2}{gray!10}{white} 
    \begin{tabular}{l | c}
    \toprule[1pt]
         Module & {Scene Flow\space(test) EPE~$\downarrow$}\\
        \midrule
        \textit{Baseline} & 0.47\\
        \textit{w / o} \textit{UG-SCA} & 0.44\\
        \textit{w / o} \textit{DAv2} & 0.46\\
        \textit{w / o} \textit{LU-KSS} & 0.49\\
        \textit{w / o} $\mathcal{L}_{ADDG}$ & 0.36 \\
        \midrule
        \textit{All} & \cellcolor{green!30}{0.35} \\
    \bottomrule[1pt]
    \end{tabular}
}
\vspace{-0.1cm}
\end{table}
\paragraph{Comparison of Different Monocular Depth Estimation Models}
As shown in \cref{ddm}, we compare CogStereo with various monocular depth models. DINOv2/3 (ViT-L) yield an EPE of 0.46, revealing limited spatial reasoning despite their strength in semantics. UniDepth improves to 0.38, benefiting from its metric depth supervision but still falling short of Depth-Anything v2, whose ViT-L variant achieves the best 0.35. These results suggest that depth-oriented models trained on large-scale, multi-scene data with semantic alignment offer stronger spatial cognition priors for CogStereo, and that increasing model capacity from ViT-S to ViT-L further enhances spatial reasoning.

\begin{table}[ht]
\vspace{-0.1cm}
\centering
\small
\renewcommand{\arraystretch}{1.0}
\caption{{Comparison of Our CogStereo Equipped with Different Monocular Depth Estimation Models}}
\label{ddm}
\resizebox*{0.9\linewidth}{!}{
    \renewcommand{\arraystretch}{1.2}
    \begin{tabular}{c| c | c}
    \toprule[1pt]
        \textbf{Model} & \textbf{Encoder} & \textbf{Scene Flow (test) EPE~$\downarrow$}\\
        \midrule
         DINOv2~\cite{oquab2023dinov2} & VIT-L & 0.46\\
         DINOv3~\cite{simeoni2025dinov3} & VIT-L & 0.46\\
        \hline
         UniDepth~\cite{piccinelli2024unidepth} & VIT-L & 0.38 \\
         \hline
         \multirow{3}*{DAv2~\cite{yang2024depth}} 
         & VIT-S & 0.42 \\ 
         & VIT-B & 0.37\\ 
         & VIT-L & \cellcolor{green!30}{0.35}\\ 
    \bottomrule[1pt]
    \end{tabular}
}
\vspace{-0.1cm}
\end{table}
\paragraph{Extension to Other Stereo Matching Framework}
To validate the generality of the proposed Spatial Cognition (SC), we integrate it into three representative stereo frameworks: RAFT-Stereo, Selective-Stereo, and IGEV. As shown in~\cref{tab:extension}, adding SC consistently improves performance: RAFT-Stereo’s EPE drops from 0.63 to 0.44, Selective-Stereo from 0.44 to 0.35, and IGEV from 0.47 to 0.35. These results demonstrate that SC is a versatile prior, effective across different stereo architectures and confirming its broad applicability and robustness.
\begin{table}[htbp]
\centering
\caption{Ablation study of the universality of proposed Spatial Cognition~(SC).}
\label{tab:extension}
\resizebox*{0.9\linewidth}{!}{
    \renewcommand{\arraystretch}{1.0}
    \begin{tabular}{c c | c}
    \toprule[1pt]
        \multirow{1}*{Model} & Module & {Scene Flow\space(test) EPE~$\downarrow$}\\
        \midrule
        \multirow{2}*{RAFT-Stereo} & \textit{w / o} SC& 0.63\\
        & \cellcolor{gray!10}\textit{w} SC& \cellcolor{green!30}0.44\\
        \midrule
        \multirow{2}*{Selective-Stereo} & \textit{w / o} SC &0.44 \\
        &\cellcolor{gray!10}\textit{w} SC & \cellcolor{green!30}0.35\\
        \midrule
        \multirow{2}*{IGEV} & \textit{w / o} SC & 0.47\\
        &\cellcolor{gray!10}\textit{w} SC & \cellcolor{green!30}0.35\\
    \bottomrule[1pt]
    \end{tabular}
}
\vspace{-0.3cm}
\end{table}
\subsection{Downstream Benchmark Performance}
As shown in the last two rows of ~\cref{fig:euroc}, in zero-shot evaluations on the EuRoC dataset for Stereo SLAM, CogStereo outperforms IGEV and Selective-IGEV by generating more coherent and accurate depth maps, especially in challenging regions with textureless, occlusions, or reflective surfaces. Notably, DAv2 produces structurally complete and semantically plausible depth maps, which underscores the feasibility of spatial cognition. By combining stereo matching with SC, CogStereo enhances depth quality without fine-tuning. Single-frame inference takes ~0.3 s on a single RTX 3090 GPU with ~19 GB memory, confirming its practical utility for autonomous driving and robotic navigation.

\section{Conclusion}

This paper introduces CogStereo to neural stereo matching that addresses ill-posed regions such as occlusions, textureless, and appearance ambiguities by embedding implicit spatial cognition. {CogStereo leverages DAv2 feature as spatial cognition}, introducing an understanding of scene layout akin to human perception, thereby enhancing the global consistency and accuracy of matching.
Comprehensive experiments across numerous standard benchmarks have shown that CogStereo has achieved top-tier performance and demonstrated remarkable generalization to ill-posed regions across diverse datasets. The success of CogStereo illustrates the potential of integrating geometric reasoning  with spatial cognition to elevate stereo matching beyond basic geometric reasoning to a more sophisticated level of cognitive understanding. 



{
\bibliographystyle{IEEEtran}
\bibliography{main}
}

\end{document}